\documentclass[letterpaper, 10 pt, conference]{ieeeconf}  % Comment this line out if you need a4paper

\IEEEoverridecommandlockouts                              % This command is only needed if 
\usepackage[ruled,vlined,linesnumbered]{algorithm2e}
\usepackage{amsmath, graphicx}
\usepackage{balance}
\usepackage{amssymb,amsbsy}
\usepackage{breqn,bbm,xcolor}%\usepackage[dvipsnames]{xcolor}
\usepackage{multirow}
\usepackage[symbol,bottom]{footmisc}
\usepackage{soul}
\usepackage{url}
\usepackage{booktabs} % To thicken table lines
\usepackage{tikz}
\usetikzlibrary{positioning}
\usetikzlibrary{arrows}
\usetikzlibrary{shapes,snakes}
\usetikzlibrary{plotmarks}
\usepackage{subcaption}
\title{\LARGE \bf
Tracking Snake-like Robots in the Wild using only a Single Camera %Mike's attempt
}
\author{Jingpei Lu$^{1}$, Florian Richter$^{1}$, Shan Lin$^{1}$, and Michael C. Yip$^{1}$ % <-this % stops a space
\thanks{$^{1}$Department of Electrical and Computer Engineering, University of California San Diego, La Jolla, CA 92093 USA.{\tt\small\{jil360, frichter, shl102, yip\}@ucsd.edu}}
}

\begin{document}

\maketitle
% \thispagestyle{empty}
% \pagestyle{empty}
%\IEEEpeerreviewmaketitle

%%%%%%%%%%%%%%%%%%%%%%%%%%%%%%%%%%%%%%%%%%%%%%%%%%%%%%%%%%%%%%%%%%%%%%%%%%%%%%%%
\begin{abstract}
Robot navigation within complex environments requires precise state estimation and localization to ensure robust and safe operations. For ambulating mobile robots like robot snakes, traditional methods for sensing require multiple embedded sensors or markers, leading to increased complexity, cost, and increased points of failure. Alternatively, deploying an external camera in the environment is very easy to do, and marker-less state estimation of the robot from this camera's images is an ideal solution: both simple and cost-effective. However, the challenge in this process is in tracking the robot under larger environments where the cameras may be moved around without extrinsic calibration, or maybe when in motion (e.g., a drone following the robot). The scenario itself presents a complex challenge: single-image reconstruction of robot poses under noisy observations. In this paper, we address the problem of tracking ambulatory mobile robots from a single camera. The method combines differentiable rendering with the Kalman filter. This synergy allows for simultaneous estimation of the robot's joint angle and pose while also providing state uncertainty which could be used later on for robust control. We demonstrate the efficacy of our approach on a snake-like robot in both stationary and non-stationary (moving) cameras, validating its performance in both structured and unstructured scenarios. The results achieved show an average error of $0.05$ m in localizing the robot's base position and $6$ degrees in joint state estimation. We believe this novel technique opens up possibilities for enhanced robot mobility and navigation in future exploratory and search-and-rescue missions.

%150--250 words.
\end{abstract}

%%%%%%%%%%%%%%%%%%%%%%%%%%%%%%%%%%%%%%%%%%%%%%%%%%%%%%%%%%%%%%%%%%%%%%%%%%%%%%%%
\section{Introduction}

Unlike their stationary counterparts, mobile robots are designed to navigate through the physical world in environments that are often too treacherous for humans such as the deep sea \cite{kunz2008deep} and even other planets \cite{squyres2003athena}.
With mobile robots acting as surrogates for humans, exploration for research and search and rescue missions in extreme environments are conducted without risking human lives \cite{whitman2018snake}.
A growing class of mobile robots involves ambulatory systems. These ambulatory mobile robots (AMRs) have specialized articulated robotic designs for enhanced mobility and stability on uneven ground techniques in order to navigate broader terrains. AMRs include but are not limited to quadruped robots \cite{bledt2018cheetah}, flying drones \cite{aabid2022reviews}, and snake-like and serpentine robots \cite{wright2007design, richter2021arcsnake}.

\begin{figure}[t]
    \vspace{0.1in}
    \centering
    \includegraphics[width=\linewidth]{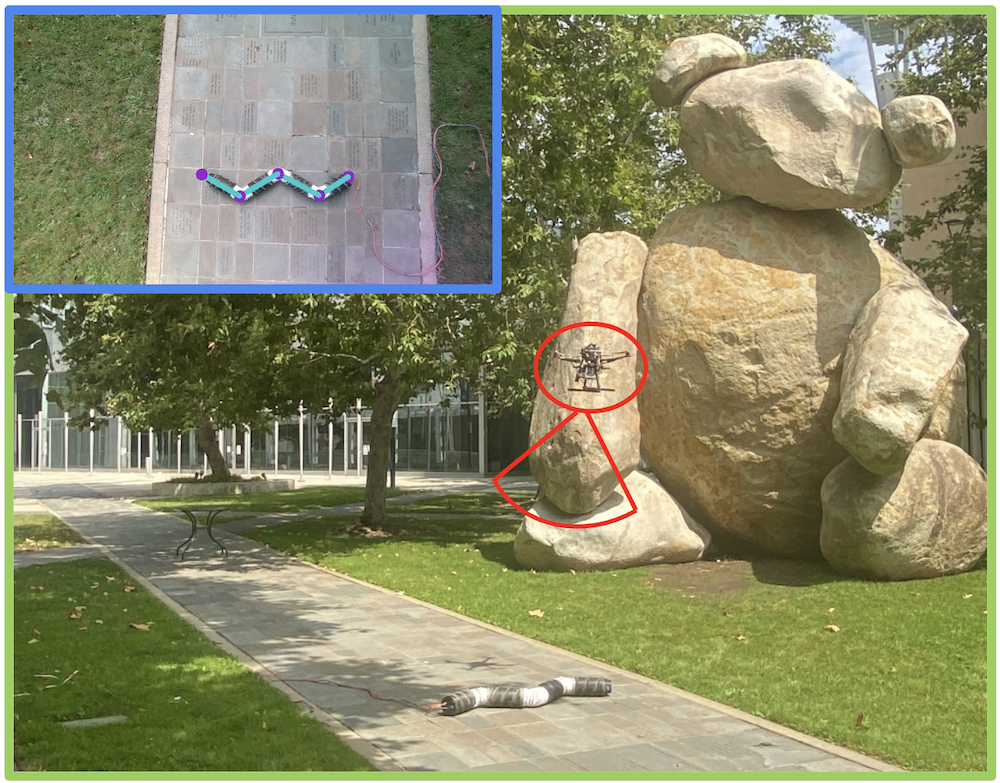}
    \caption{A snake-like robot, Arcsnake~\cite{richter2021arcsnake}, is tracked on camera in the outdoor environment by a hovering drone. }
    \label{fig:cover}
\end{figure}

To ensure the safe operation of AMRs in complex environments, various sensors are integrated into their systems. These sensors aid in localizing the robot and understanding its surroundings, though this can introduce increased complexity in real-world deployments.
A more streamlined approach involves tracking AMRs using cameras. Cameras, given their ease of installation and portability, are better for navigating challenging terrains. 
For example, in the Mars 2020 NASA mission, where the Mars Helicopter utilized onboard cameras to scout the landscape and guide the Perseverance rover's exploration. 
As we look to the future, exploratory and search-and-rescue missions likely involve collaborative efforts between multiple robots, and the ability to track one robot using a camera mounted on another will be crucial.

%Due to their high maneuverability and on-board sensing, drones have a significant number of applications in surveillance such as traffic monitoring, freight delivery, road construction and photogrammetry \cite{barmpounakis2016unmanned}.
%Algorithms have even been presented to conduct human pose estimation \cite{pirinen2019domes} and reconstruct buildings in 3D \cite{daftry2015building} from video data captured by drones.
%During the Mars 2020 NASA mission, the Perseverance rover utilized Ingenuity Mars Helicopter to preview areas of possible interest.
% NASA's Ingenuity Mars Helicopter previewed areas of possible interest for the Perseverance rover to explore.

In this paper, we address the problem of tracking snake-like robots from a single camera.
Along the lines of the Mars Helicopter's mission, we aim to bring robot state estimation from camera data to snake-like robots, and by extension, other AMRs, to aid in future exploratory missions.
By estimating the pose and state of an AMR, drones can provide more detailed guidance when providing mapping of the environment \cite{von2017monocular}.
Our focus is on snake robots that draw inspiration from biological snakes \cite{pettersen2017snake} and are currently funded by NASA for exploration on extraterrestrial planetary bodies \cite{carpenter2021exobiology}.
Toward this end, we recognize a fundamental need for being able to track AMRs using only a monocular camera. These techniques will also become foundational in the future to deploying robots in search-and-rescue missions or leveraging autonomous robot teams for work in the remote wilderness.

The overall tracking approach involves first a method for automatic robot mask generation. Leveraging this mask, we present a tracking technique that seamlessly integrates differentiable rendering with the Kalman filter, ensuring precise online state estimation. %The proposed algorithm first employs image moments \cite{chaumette2004image} to obtain a rough estimation with belief propagation. Then, the robot state is refined by rendering this estimation and minimizing the discrepancy with the actual image observation. 
We conduct experiments in both laboratory and outdoor environments (Figure \ref{fig:cover}). Through both qualitative and quantitative evaluations, we demonstrate the effectiveness of our method in different scenarios. Our contributions are threefold:
\begin{itemize}
    \item We present the first work on marker-less state estimation for a snake robot from a single monocular camera.
    \item Our method combines differentiable rendering with a Kalman filter, and simultaneously estimates the joint angle and the pose of a snake robot.
    \item Validation of the effectiveness of the algorithm on a snake robot in both structured and unstructured environments, achieving a localization accuracy of 0.05 m for the robot base position and 0.11 rad on the robot's joint states.
\end{itemize}

\section{Previous Work}

\subsection{Robot Localization from Single Camera}

Localizing the robot is crucial for a wide range of robotic applications, especially when relying on a single camera, which presents unique challenges. One popular approach to address this is using the fiducial markers as 2D point features \cite{garrido2014aruco,olson2011apriltag}. 
For articulated robots like a snake robot, the 3D position of the markers can be calculated using robot kinematics and the robot pose can be derived by solving a Perspective-n-Point problem~\cite{park_robot_1994,fassi2005hand,ilonen2011robust,horaud1995hand}.

As the field evolved, there was a shift towards marker-less pose estimation. Initial efforts in this direction utilized depth cameras to localize articulated robots \cite{schmidt2014dart,pauwels2014real,michel2015pose,desingh2019factored}. With the rise of Deep Neural Networks (DNNs), a new paradigm emerged. DNNs, with their advantages of extracting point features without the need for markers, have significantly enhanced the performance of marker-less pose estimation for articulated robots \cite{lambrecht2019towards,lee2020dream,lu2022keypoint,zuo2019craves,lu2023markerless}. Beyond keypoint-based methods, recent works \cite{labbe2021robopose,lu2023} have demonstrated the potential of rendering-based methods. Benefiting from the dense correspondence provided by robot masks, rendering-based methods achieve state-of-the-art performance on robot pose estimation. However, they suffer from processing speed.

In this work, we adopt a rendering-based approach for robot state estimation. Instead of purely relying on the rendering, we integrate image moments with a Kalman Filter, aiming to utilize temporal information to achieve precise and fast online inference using a single camera.

\subsection{Snake Robot State Estimation}

%State estimation plays a critical role in the operation of mobile robots, enabling them to accurately perceive and navigate their environment. For mobile robots, the primary focus of state estimation is on localizing the robot within its surroundings. For instance, Milella et al. \cite{milella2006computer} utilizes visually distinctive features on stereo images for localization. Several other works \cite{goldberg2003maximizing, zouaghi2011probabilistic, colle2019robust} have proposed methods that take into account the dynamic nature of the environment and potential measurement errors to enhance localization accuracy.
For a broader category of mobile robots, the primary focus of state estimation has been on localizing the robot within its surroundings. For instance, Milella et al. \cite{milella2006computer} utilizes visually distinctive features on stereo images for localization. Several other works \cite{goldberg2003maximizing, zouaghi2011probabilistic, colle2019robust} have proposed methods that take into account the environment dynamics and potential measurement errors to enhance localization accuracy.

However, in the realm of snake robots, state estimation becomes even more intricate due to the need to consider joint angles for accurate 3D space modeling. Historically, state estimation for snake robots has relied on the robot's internal proprioceptive sensors, as highlighted by works like Rollinson et al. \cite{rollinson2011state,rollinson2013robust}. Then, the filtering methods, like the Unscented
Kalman Filter and Extended Kalman Filter \cite{kalman1961new,van2004sigma}, have been employed to account for the measurement error for real-time estimation.

In this work, we seek to estimate both the position and joint angle of the snake robot using only images. This approach not only simplifies the estimation process but also enhances the robot's adaptability in outdoor scenarios.

\begin{algorithm}[t]
    \caption{Online State Estimation}
    \label{alg:main_outline_2}
    %\small
    \SetKwInOut{Input}{Input}
    \SetKwInOut{Output}{Output}
    \Input{Initialized robot state $\textbf{x}_{0|0},\Sigma_{0|0}$}
    \Output{Estimated robot state $\textbf{x}_{t|t}, \Sigma_{t|t}$}
    % \tcp{The state x =: [joint angles, camera to robot transform]}
    \While{receive new image $\mathbb{I}_t$}
    {
        \tcp{Motion Model}
        $\textbf{x}_{t|t-1},\Sigma_{t|t-1} \leftarrow motionModel( \textbf{x}_{t-1|t-1}, \textbf{v}_{t-1},\Sigma_{t-1|t-1})$ \\
        \tcp{Observation from Image}
        $\mathbb{M}^{ref}_t \leftarrow f_{seg}(\mathbb{I}_t)$\\
        $ \textbf{m}_t \leftarrow computeMoments(\mathbb{M}^{ref}_t)$ \\
        
        \tcp{Observation Model}
        $\mathcal{M}_{t|t-1} \leftarrow reconstructMesh(\textbf{x}_{t|t-1})$\\
        $\mathbb{M}^{pred}_{t|t-1} \leftarrow renderPrediction(\textbf{x}_{t|t-1}, \mathcal{M}_{t|t-1})$\\ 
        $ \hat{\textbf{m}}_t \leftarrow computeMoments(\mathbb{M}^{pred}_{t|t-1})$ \\
        $H_t = \frac{\partial \hat{\textbf{m}}_t}{\partial \textbf{x}_{t|t-1}}$\\
        
        \tcp{Compute the Residual}
        $\textbf{y}_t = \textbf{m}_t - \hat{\textbf{m}}_t$

        \tcp{Update Belief}
        $K_t = \Sigma_{t|t-1} H_t^{\top} (H_t \Sigma_{t|t-1}H_t^{\top})^{-1}$\\
        $\textbf{x}_{t|t} = \textbf{x}_{t|t-1} + K_t \textbf{y}_t$\\
        $\Sigma_{t|t} = (I - K_t H_t) \Sigma_{t|t-1}$\\
        
        \tcp{Refine with Image Loss}
        \For{number of refinement steps}{
        $\mathcal{M}_{t|t} \leftarrow reconstructMesh(\textbf{x}_{t|t})$\\
        $\mathbb{M}^{pred}_{t|t} \leftarrow renderPrediction(\textbf{x}_{t|t},\mathcal{M}_{t|t})$\\
        $\mathcal{L}_t \leftarrow computeLoss(\mathbb{M}^{pred}_{t|t}, \mathbb{M}^{ref}_t)$\\
        $\textbf{x}_{t|t} = \textbf{x}_{t|t} - \lambda \frac{\partial \mathcal{L}_t}{\partial \textbf{x}_{t|t}}$\\
        }
        
        \tcp{Update Velocity}
        $\textbf{v}_t \leftarrow computeVelocity(\textbf{x}_{t|t}, \textbf{x}_{t-1|t-1})$
    }
\end{algorithm}

\section{Methodology}

The overall proposed approach follows an online state estimation method combining differentiable rendering of a robot mask, with image moment prediction, a robot motion model, and a Kalman filter to estimate the joint angle and the pose of a mobile robot from a single camera. The method includes, additionally, refinement steps and velocity update steps to enhance the accuracy of the estimation, as well as model transfer techniques to reduce computation and memory costs so that the method can run on modest hardware. The details follow in the next section, and Algorithm \ref{alg:main_outline_2} outlines the main steps of the method.

\subsection{Motion Model with Belief Propagation}
\label{sec:motion_modl}

For AMR navigation, the robot state, denoted by $\mathbf{x}_t$, can encapsulate various attributes such as joint angles, camera-to-robot transformations, and other necessary parameters at time $t$. In this work, we define the robot state as $\mathbf{x}:= [\theta, \mathbf{q},\mathbf{b}]$, where $\theta \in \mathbb{R}^N$ is the robot joint angle, $\mathbf{q}$ is the quaternion, and $\mathbf{b}$ is the translational vector. 
The quaternion and the translational vector are parametrizations of the $\mathbf{T}^c_b \in SE(3)$, which is the robot pose in the camera frame.

The next state of the robot is predicted with a motion model, based on its previous state and velocity. This prediction phase provides a rough direction for belief propagation. We will model the robot's motion using a simple linear relationship:
\begin{equation}
    \textbf{b}_{t|t-1} = \textbf{b}_{t-1|t-1} + \textbf{v}_{t-1} \Delta t
\end{equation}
where we try to predict the position of the robot $\textbf{b}_{t|t-1}$ at time $t$ by considering the previous robot position $\textbf{b}_{t-1|t-1}$, the velocity $\textbf{v}_{t-1}$, and the time step $\Delta t$. We will make the assumption that there is negligible process noise (i.e., imperfections in the system's motion model are negligible as compared to observation noise), leading to the following expression for the propagation of the covariance matrix:
\begin{equation}
    \Sigma_{t|t-1} = F_t \Sigma_{t-1|t-1} F_t^{\top}
\end{equation}
In this case, $F_t$ is the identity matrix, reflecting our assumption that the motion model follows a linear relationship without any non-linear or stochastic effects.

\subsection{Automatic Mask Generation for Segmentation}
\label{sec:segmentation}

The proposed state estimation algorithm requires segmenting the robot from images, but manually labeling the robot masks can be highly time-consuming. Recently, the zero-shot generalizable segmentation model, Segment Anything Model (SAM) \cite{kirillov2023segment}, allows automatic robot mask generation with simple bounding box prompts. 

\begin{figure}[t!]
\vspace{0.1in}
\centering
\includegraphics[width=0.85\linewidth]{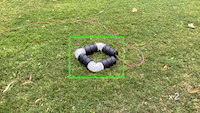}
\includegraphics[width=0.85\linewidth]{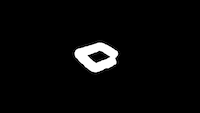}
\caption{Example of the bounding box prompt generated by mask-to-box operation (top) and the corresponding robot mask generated using SAM (bottom).}
\label{fig:bbox}
%\vspace{-0.14in}
\end{figure}

Given the binary robot mask of the previous frame, $\mathbb{M}_{t-1} \in \mathbb{R}^{H \times W}$, the bounding box prompt for the current frame, $\mathcal{B}_t:=(u_{min},v_{min}, u_{max}, v_{max})$, is estimated by a mask-to-box operation,
\begin{align}
(u_{min}, v_{min}) & = \min \{ (u,v) \,|\, \mathbb{M}_{t-1}[u,v] \neq 0 \} \\
(u_{max}, v_{max}) & = \max \{ (u,v) \,|\, \mathbb{M}_{t-1}[u,v] \neq 0 \} 
\end{align}
Then, the SAM is utilized to generate the robot mask of the current frame, given the bounding box prompt $\mathcal{B}_t$, as shown in Fig. \ref{fig:bbox}. 
To ensure the robustness of the bounding box prompt, the robot mask is dilated before performing the mask-to-box operation.

Using SAM for robot mask generation can, however, be slow as SAM is not optimized for real-time application (around 0.5 seconds per frame using a single Nvidia GeForce RTX 4090 GPU). 
To achieve real-time performance, we utilize the robot masks generated from SAM to train a lightweight neural network for segmentation. Specifically, we employ DeepLabV3+ \cite{chen2018encoder}, a popular semantic segmentation architecture, to segment the robot from RGB images during the online estimation process.
By training DeepLabV3+ with the generated masks, we ensure that our system can segment the robot in real-time with modest memory and computation requirements, effectively enabling realistic deployment in the wild.

\subsection{Observation Model for Belief Propagation}
\label{sec:observation_model}

%Updating the robot states with image observation is crucial for facilitating belief propagation in the proposed algorithm. The link between the predicted states and the observation serves as the observation model for the Kalman filter. 
In this section, we introduce the mapping from the predicted robot states $\textbf{x}_{t|t-1}$ to the observation of image moment \cite{chaumette2004image} $\hat{\textbf{m}}_t$ in the proposed algorithm \ref{alg:main_outline_2}.

Given the predicted robot states $\textbf{x}_{t|t-1}$, which includes joint angle and robot pose, we first reconstruct the robot mesh by interconnecting individual robot body parts through forward kinematics. 
For a snake-like (serpentine) robot, we approximate each individual robot body part as a cylinder with the dimension mentioned in \cite{schreiber2020arcsnake, richter2021arcsnake}.
Given a mesh vertex $\mathbf{r}^n \in \mathbb{R}^3$ on the 
$n$-th robot link, this vertex undergoes a transformation into the robot base frame considering the joint angle:
\begin{equation}
\label{eq:vertex_transform}
    \overline{\mathbf{r}}^b = \mathbf{T}^b_n(\theta) \overline{\mathbf{r}}^n
\end{equation}
where $\overline{\cdot}$ represents the homogeneous representation of a point (i.e. $\overline{\mathbf{r}} = [\mathbf{r}, 1]^T$), and $\mathbf{T}^b_n(\theta)$ is the coordinate frame transformation obtained from the forward kinematics~\cite{denavit1955kinematic}.

Having the reconstructed robot mesh and the predicted robot base-to-camera transformation, $\mathbf{T}^c_b$, the PyTorch3D differentiable renderer \cite{pytorch3d} comes into play to produce a virtual-model-derived, or rendered robot mask. By referencing techniques similar to those in \cite{lu2023}, a differentiable silhouette renderer paired with a perspective camera is employed. The \textit{SoftSilhouetteShader} is specifically leveraged to compute pixel values that form the robot mask.

With the rendered robot mask, $\mathbb{M}$, the image moments become computable as:
\begin{equation}
    M_{ij} = \sum_u \sum_v u^i v^j \mathbb{M}(u,v)
\end{equation}
Then, we derive the centroid, which is our observation for belief propagation, by:
\begin{equation}
    \hat{\textbf{m}} = \begin{bmatrix}\frac{M_{10}}{M_{00}} & \frac{M_{01}}{M_{00}} \end{bmatrix}^\top
\end{equation}
We employ pytorch autograd \cite{paszke2017automatic} to track the gradient of each step and compute the observation matrix $H$ by collecting the derivatives of the image moment $\hat{\textbf{m}}$ with respect to the robot states $\textbf{x}_{t|t-1}$.

Finally, an Extended Kalman Filter (EKF) \cite{kalman1961new} is employed to update the belief of the robot states (lines 9-12 in Alg. \ref{alg:main_outline_2}), which ensures that our belief about the robot states is continually refined as more observations come in.

\subsection{Image Loss Refinement and Velocity Estimation}
\label{sec:refinement}

While image moments have historically proven useful in object tracking \cite{chaumette2004image,yao2020image}, their efficacy diminishes in the complex arena of robot state estimation. This is because they encapsulate only limited details of the robot mask. Consequently, a direct method that compares the estimated and reference robot masks provides an enhancement to state estimation accuracy.

We predict the robot mask from estimated robot states using the same differentiable rendering pipeline as described in Section \ref{sec:observation_model}. To measure the difference between this prediction and the reference mask, we employ an image loss function, which sums the squared differences between the predicted mask $\mathbb{M}^{pred}$ and the reference mask $\mathbb{M}^{ref}$ across the image dimensions:
\begin{equation}
        \mathcal{L} = \sum_{i=0}^{H-1} \sum_{j=0}^{W-1} \left(\mathbb{M}^{pred}(i,j) - \mathbb{M}^{ref}(i,j)\right)^2.
\label{eq:mask_loss}
\end{equation}
We refine the mean of the robot states by applying back-propagation on this image loss (line 17 in Alg. \ref{alg:main_outline_2}), bringing the estimation closer to the true state.

As a final step, in service of the next belief propagation timestep, we derive the velocity from the updated position:
\begin{equation}
    \textbf{v}_t = \frac{\mathbf{b}_{t|t} - \mathbf{b}_{t-1|t-1}}{\Delta t}
\end{equation}
This velocity is used for the motion model in forthcoming iterations, as it feeds into predictions for the robot's future states.

\section{Experiments and Results}
\label{section:experiment}

To comprehensively assess the efficacy of our proposed state estimation algorithm, we collected datasets of a snake robot operating in both structured and unstructured environments. These datasets facilitated both qualitative and quantitative evaluations of the state estimation method.

The snake robot hardware is described in \cite{schreiber2020arcsnake,richter2021arcsnake} and is the evolutionary precursor to the NASA Extant Exobiology Life Surveyor (EELS) robot \cite{carpenter2021exobiology} that is anticipated to serve a science research vehicle for both earth science missions as well as extraterrestrial planetary exploration on Saturn's moon, Enceladus, or Jupiter's moon, Europa.

\textbf{Snake-Lab Dataset}:
We introduced the Snake-Lab Dataset for evaluating the accuracy of the joint angle estimation and robot pose estimation.
This dataset was acquired in a lab setting using an Intel\textregistered{} Realsense\texttrademark{} camera at a resolution of (1280, 720). The robot's joint angles were recorded using electromagnetic sensors and were synchronized with the captured images. Additionally, the robot's spatial position was determined using the depth capabilities of the camera. For evaluation metrics, we employed the Euclidean distance for position estimation and the $L_1$ norm for joint angle estimation.

\textbf{Snake-Outdoor Dataset}: To examine the robustness of our algorithm in less structured environments, we collected the Snake-Outdoor dataset. This dataset comprises three videos: the first two were recorded using a hand-held camera at a resolution of (1280, 720), while the third was captured via a drone camera, which has no direct connection to the snake robot system. Given the absence of ground truth for the robot's state in this setting, we adopted the Intersection-over-Union metric (IoU):
\begin{equation}
    \text{IoU} = \frac{|\mathbb{M}^{ref} \cap \mathbb{M}^{pred}|}{|\mathbb{M}^{ref} \cup \mathbb{M}^{pred}|} 
\end{equation}
to compare the ground-truth robot mask $\mathbb{M}^{ref}$ with our algorithm's estimated mask $\mathbb{M}^{pred}$.

\begin{figure}[t]
    \vspace{0.1in}
    \centering
    \includegraphics[width=\linewidth]{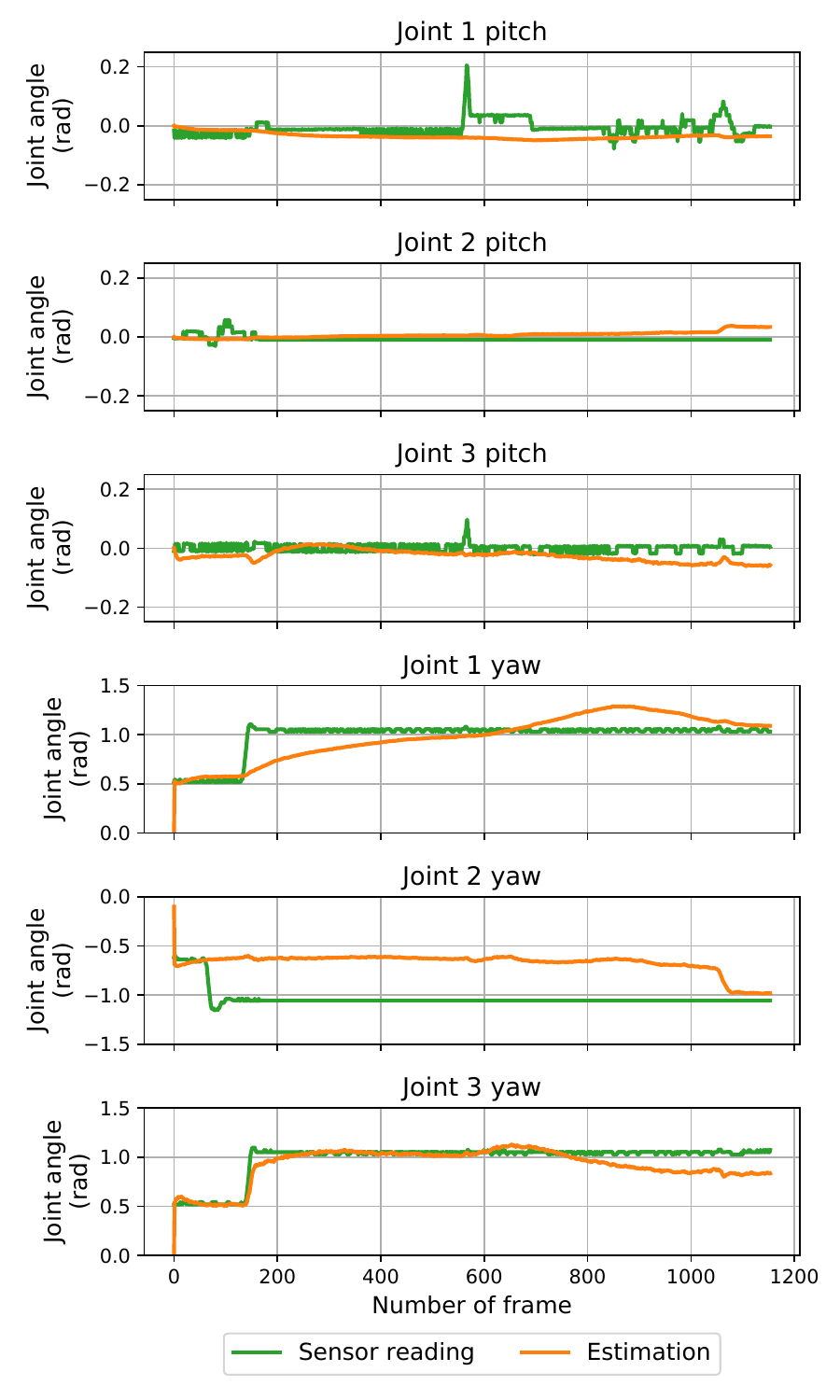}
    \caption{Plots of estimated joint trajectory vs. sensor reading for the Snake-Lab dataset in the moving robot scenario. For each joint, we plot the pitch and yaw angle separately. Note that the snake robot uses magnetic encoders for sensor readings and are slightly noisy due misalignment between the encoder and magnet from vibrations during the experiment.}
    \label{fig:joint_angle}
\end{figure}

\begin{figure}[t]
    \vspace{0.1in}
    \centering
    \includegraphics[width=\linewidth]{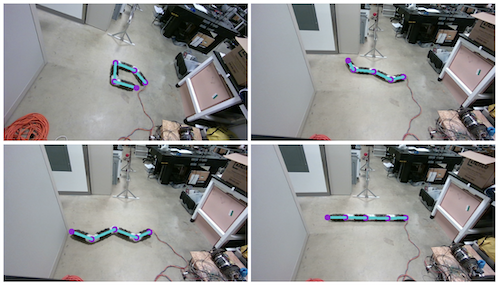}
    \caption{Qualitative results on Snake-Lab dataset. We derive the skeleton from the estimated robot pose and joint angle, and visualize it by projecting the skeleton on images. }
    \label{fig:lab}
\end{figure}

\begin{table}
\centering
\caption{Average Position and State Estimation Error on Snake-Lab Dataset}
\begin{tabular}{lcc}
\toprule
  & Position error (m) & Joint state error (rad)\\
\midrule
Static & 0.0278  &  0.0605 \\
Moving camera &  0.0647 & 0.0849 \\
Moving robot &  0.0587 & 0.1352 \\
\midrule
Overall &  0.0540 & 0.1125 \\
\bottomrule
\end{tabular}
\label{tab:lab}
\end{table}

\subsection{Implementation Details}

To train DeepLabV3+, we collected around 1500 images, captured at a resolution of (1280, 720) and the ground truth segmentation masks were generated using Segment Anything Model \cite{kirillov2023segment}.
We used the Adam optimizer \cite{kingma2014adam} for gradient descent with 20 epochs and 8 batch size. The initial learning rate was set to 0.0001 and was decayed by a factor of 0.1 at the 10th epoch. 

During the online estimation, we resize the raw image to a resolution of (640, 360). Both the observed robot mask and the rendered robot mask are processed at this resolution. For the refinement step, we set the learning rate to 0.005 and also used the Adam optimizer for gradient descent.
All computational experiments were executed on a system equipped with an Intel\textregistered{} Core\texttrademark{} i9-11900F Processor and NVIDIA GeForce RTX 4090.
To strike a balance between accuracy and processing speed, we perform 10 refinement iterations for each incoming image, ensuring optimal performance while sustaining an estimation speed of 1 FPS.

\subsection{Experiment on Snake-Lab dataset}

We present the qualitative results on the Snake-Lab dataset in Figure \ref{fig:lab}, and the quantitative evaluation of our state estimation algorithm is presented in Table \ref{tab:lab}. We also plot the estimated joint trajectory with sensor readings in Figure \ref{fig:joint_angle}. 

The results are segmented based on different scenarios: static conditions, moving camera, and moving robot. 
Under static conditions, where both the camera and the robot remain stationary, both the joint angle error and position error are the lowest, indicating that the algorithm performs exceptionally well in stable environments.
Moving the camera or robot slightly affects the algorithm's accuracy.
This could be attributed to the dynamic nature of the camera and the robot's movements, which might introduce complexities in state estimation.
The overall average position error and joint angle error across all scenarios are 0.0540 m and 0.1125 rad, respectively. These results affirm the robustness of our state estimation algorithm, even in varying conditions. However, it's evident that dynamic factors, such as camera or robot movement, introduce some challenges, leading to increased errors.

\begin{table}[b]
\centering
\caption{Quantitative evaluation on Snake-Outdoor dataset. We compute the IoU between the estimated robot mask and the ground-truth robot mask. We also report the processing speed under different settings.}
\begin{tabular}{lccc}
\toprule
 & \multicolumn{3}{c}{Number of refinement steps}\\
\cmidrule(lr){2-4}
  & 1 & 5 & 10\\
\midrule
Video 1 (Mean IoU)  & 0.4659 & 0.7632  &  0.8665 \\
Video 2 (Mean IoU) & 0.2456 & 0.3584 &  0.7690 \\
Video 3 (Mean IoU)& 0.3088 & 0.4394 &  0.8210 \\
\midrule
Speed (FPS)  & 3.5 & 1.5  & 1 \\
\bottomrule
\end{tabular}
\label{tab:outdoor}
\end{table}

\subsection{Experiment on Snake-Outdoor dataset}

\begin{figure*}[t]
    %\vspace{0.1in}
    \centering
    \includegraphics[width=.98\linewidth]{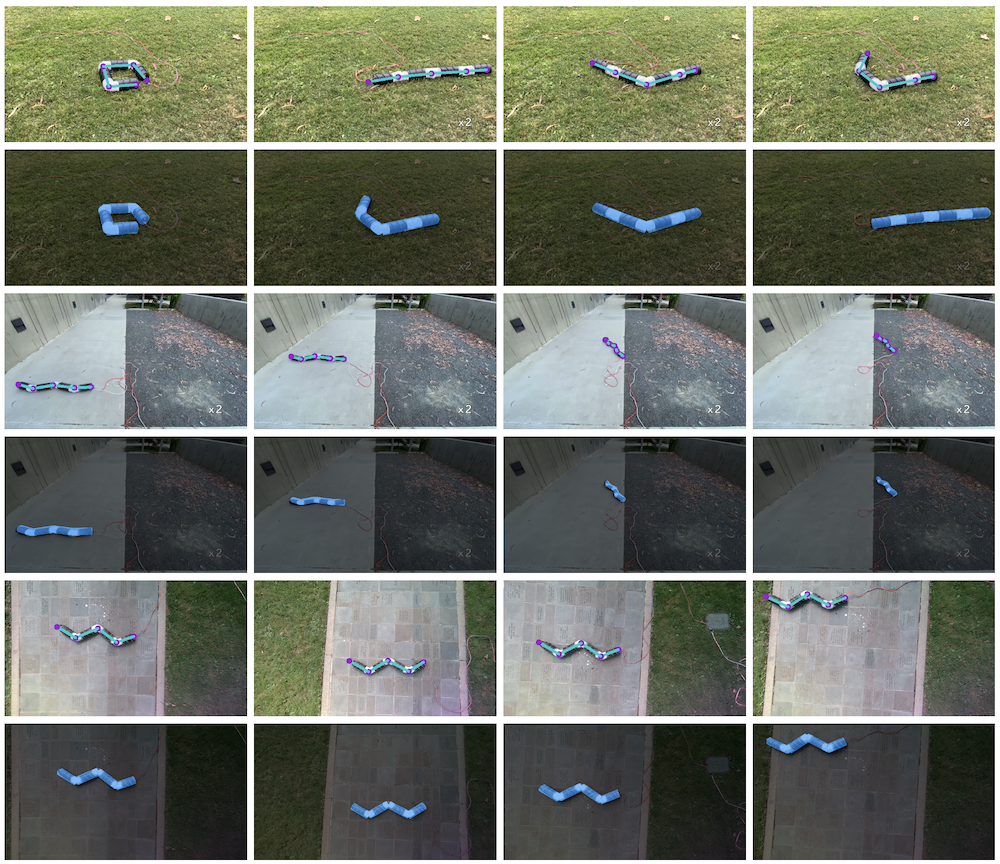}
    \caption{Qualitative results on Snake-Outdoor dataset. We show the estimated skeleton and predicted robot mask overlaid on images. Rows 1-2 correspond to video 1, rows 3-4 correspond to video 2, and rows 5-6 correspond to video 3. Notably, there's a precise alignment of the skeleton and mask with the robot as shown in the images.}
    \label{fig:outdoor}
\end{figure*}

Table \ref{tab:outdoor} presents the quantitative evaluation of our state estimation algorithm on the Snake-Outdoor dataset. The results are organized based on the number of refinement steps taken, which are 1, 5, and 10. The performance metric used is the Intersection-over-Union (IoU) for each video, and the speed of the algorithm in frames per second (FPS) is also provided.
From the Table \ref{tab:outdoor}, we can see a clear trade-off between accuracy and speed. As the number of refinement steps increases, there is a noticeable improvement in the Mean IoU, but the speed decreases. With 10 refinement steps, the algorithm operates at 1 FPS, which might be a limiting factor for real-time applications. However, the significant boost in accuracy might justify this trade-off in scenarios where precision is critical.

We also present qualitative results in Fig \ref{fig:outdoor}, showing the estimated skeleton and the predicted robot mask overlaid on the images. 
We can observe the estimated skeleton aligns with the robot's actual structure, providing a clear and intuitive understanding of the algorithm's performance in real-world, outdoor settings.

\section{Conclusion}

In this work, we present a novel method for state estimation of snake robots using a single camera. The proposed approach combines differentiable rendering with the Kalman filter, fusing temporal information with a rendering-based optimization technique to improve the estimation process, which enhances the method's adaptability in outdoor scenarios. The results demonstrate the efficacy of our approach on a snake robot, validating its performance in both structured and unstructured environments. We believe this technique opens up possibilities for expanded capabilities for ambulatory mobile robot deployment and navigation in complex environments, making it a promising solution for future mobile robot applications.

For future works, an exciting avenue is the exploration of how our method can be adapted for collaborative robotics, where multiple robots work in tandem. This could involve state estimation in scenarios where robots share sensory data to navigate or perform tasks (e.g. drone-assisted routing in different landscapes).

\section*{Acknowledgement}

We thank Professor Nikolay Atanasov and Jason Stanley from the Existential Robotics Laboratory at UCSD for his assistance with the drone experiments, and NASA Jet Propulsion Laboratory for their continued mission guidance.

\balance
\bibliographystyle{ieeetr}
\bibliography{references}

\begin{thebibliography}{10}

\bibitem{kunz2008deep}
C.~Kunz, C.~Murphy, R.~Camilli, H.~Singh, J.~Bailey, R.~Eustice, M.~Jakuba,
  K.-i. Nakamura, C.~Roman, T.~Sato, {\em et~al.}, ``Deep sea underwater
  robotic exploration in the ice-covered arctic ocean with auvs,'' in {\em 2008
  IEEE/RSJ International Conference on Intelligent Robots and Systems},
  pp.~3654--3660, IEEE, 2008.

\bibitem{squyres2003athena}
S.~W. Squyres, R.~E. Arvidson, E.~T. Baumgartner, J.~F. Bell~III, P.~R.
  Christensen, S.~Gorevan, K.~E. Herkenhoff, G.~Klingelh{\"o}fer, M.~B. Madsen,
  R.~V. Morris, {\em et~al.}, ``Athena mars rover science investigation,'' {\em
  Journal of Geophysical Research: Planets}, vol.~108, no.~E12, 2003.

\bibitem{whitman2018snake}
J.~Whitman, N.~Zevallos, M.~Travers, and H.~Choset, ``Snake robot urban search
  after the 2017 mexico city earthquake,'' in {\em 2018 IEEE international
  symposium on safety, security, and rescue robotics (SSRR)}, pp.~1--6, IEEE,
  2018.

\bibitem{bledt2018cheetah}
G.~Bledt, M.~J. Powell, B.~Katz, J.~Di~Carlo, P.~M. Wensing, and S.~Kim, ``Mit
  cheetah 3: Design and control of a robust, dynamic quadruped robot,'' in {\em
  2018 IEEE/RSJ International Conference on Intelligent Robots and Systems
  (IROS)}, pp.~2245--2252, IEEE, 2018.

\bibitem{aabid2022reviews}
A.~Aabid, B.~Parveez, N.~Parveen, S.~A. Khan, J.~Zayan, and O.~Shabbir,
  ``Reviews on design and development of unmanned aerial vehicle (drone) for
  different applications,'' {\em J. Mech. Eng. Res. Dev}, vol.~45, no.~2,
  pp.~53--69, 2022.

\bibitem{wright2007design}
C.~Wright, A.~Johnson, A.~Peck, Z.~McCord, A.~Naaktgeboren, P.~Gianfortoni,
  M.~Gonzalez-Rivero, R.~Hatton, and H.~Choset, ``Design of a modular snake
  robot,'' in {\em 2007 IEEE/RSJ International Conference on Intelligent Robots
  and Systems}, pp.~2609--2614, IEEE, 2007.

\bibitem{richter2021arcsnake}
F.~Richter, P.~V. Gavrilov, H.~M. Lam, A.~Degani, and M.~C. Yip, ``Arcsnake:
  Reconfigurable snakelike robot with archimedean screw propulsion for
  multidomain mobility,'' {\em IEEE Transactions on Robotics}, vol.~38, no.~2,
  pp.~797--809, 2021.

\bibitem{von2017monocular}
L.~von Stumberg, V.~Usenko, J.~Engel, J.~St{\"u}ckler, and D.~Cremers, ``From
  monocular slam to autonomous drone exploration,'' in {\em 2017 European
  Conference on Mobile Robots (ECMR)}, pp.~1--8, IEEE, 2017.

\bibitem{pettersen2017snake}
K.~Y. Pettersen, ``Snake robots,'' {\em Annual Reviews in Control}, vol.~44,
  pp.~19--44, 2017.

\bibitem{carpenter2021exobiology}
K.~Carpenter, A.~Thoesen, D.~Mick, J.~Martia, M.~Cable, K.~Mitchell,
  S.~Hovsepian, J.~Jasper, N.~Georgiev, R.~Thakker, A.~Kourchians, B.~Wilcox,
  M.~Yip, and H.~Marvi, {\em Exobiology Extant Life Surveyor (EELS)},
  pp.~328--338.
\newblock ASCE Library, 2021.

\bibitem{garrido2014aruco}
S.~Garrido-Jurado {\em et~al.}, ``Automatic generation and detection of highly
  reliable fiducial markers under occlusion,'' {\em Pattern Recognition},
  vol.~47, no.~6, pp.~2280--2292, 2014.

\bibitem{olson2011apriltag}
E.~Olson, ``Apriltag: A robust and flexible visual fiducial system,'' in {\em
  2011 IEEE international conference on robotics and automation},
  pp.~3400--3407, IEEE, 2011.

\bibitem{park_robot_1994}
F.~C. Park and B.~J. Martin, ``Robot sensor calibration: solving {AX}= {XB} on
  the {Euclidean} group,'' {\em IEEE Transactions on Robotics and Automation},
  vol.~10, no.~5, pp.~717--721, 1994.

\bibitem{fassi2005hand}
I.~Fassi and G.~Legnani, ``Hand to sensor calibration: A geometrical
  interpretation of the matrix equation ax= xb,'' {\em Journal of Robotic
  Systems}, vol.~22, no.~9, pp.~497--506, 2005.

\bibitem{ilonen2011robust}
J.~Ilonen and V.~Kyrki, ``Robust robot-camera calibration,'' in {\em 2011 15th
  International Conference on Advanced Robotics (ICAR)}, pp.~67--74, IEEE,
  2011.

\bibitem{horaud1995hand}
R.~Horaud and F.~Dornaika, ``Hand-eye calibration,'' {\em The international
  journal of robotics research}, vol.~14, no.~3, pp.~195--210, 1995.

\bibitem{schmidt2014dart}
T.~Schmidt, R.~A. Newcombe, and D.~Fox, ``Dart: Dense articulated real-time
  tracking.,'' in {\em Robotics: Science and Systems}, vol.~2, pp.~1--9,
  Berkeley, CA, 2014.

\bibitem{pauwels2014real}
K.~Pauwels, L.~Rubio, and E.~Ros, ``Real-time model-based articulated object
  pose detection and tracking with variable rigidity constraints,'' in {\em
  Proceedings of the IEEE Conference on Computer Vision and Pattern
  Recognition}, pp.~3994--4001, 2014.

\bibitem{michel2015pose}
F.~Michel, A.~Krull, E.~Brachmann, M.~Y. Yang, S.~Gumhold, and C.~Rother,
  ``Pose estimation of kinematic chain instances via object coordinate
  regression.,'' in {\em BMVC}, pp.~181--1, 2015.

\bibitem{desingh2019factored}
K.~Desingh, S.~Lu, A.~Opipari, and O.~C. Jenkins, ``Factored pose estimation of
  articulated objects using efficient nonparametric belief propagation,'' in
  {\em 2019 International Conference on Robotics and Automation (ICRA)},
  pp.~7221--7227, IEEE, 2019.

\bibitem{lambrecht2019towards}
J.~Lambrecht and L.~K{\"a}stner, ``Towards the usage of synthetic data for
  marker-less pose estimation of articulated robots in rgb images,'' in {\em
  2019 19th International Conference on Advanced Robotics (ICAR)},
  pp.~240--247, IEEE, 2019.

\bibitem{lee2020dream}
T.~E. Lee, J.~Tremblay, T.~To, J.~Cheng, T.~Mosier, O.~Kroemer, D.~Fox, and
  S.~Birchfield, ``Camera-to-robot pose estimation from a single image,'' in
  {\em 2020 IEEE International Conference on Robotics and Automation (ICRA)},
  pp.~9426--9432, IEEE, 2020.

\bibitem{lu2022keypoint}
J.~Lu, F.~Richter, and M.~C. Yip, ``Pose estimation for robot manipulators via
  keypoint optimization and sim-to-real transfer,'' {\em IEEE Robotics and
  Automation Letters}, vol.~7, no.~2, pp.~4622--4629, 2022.

\bibitem{zuo2019craves}
Y.~Zuo, W.~Qiu, L.~Xie, F.~Zhong, Y.~Wang, and A.~L. Yuille, ``Craves:
  Controlling robotic arm with a vision-based economic system,'' in {\em
  Proceedings of the IEEE/CVF Conference on Computer Vision and Pattern
  Recognition}, pp.~4214--4223, 2019.

\bibitem{lu2023markerless}
J.~Lu, F.~Richter, and M.~C. Yip, ``Markerless camera-to-robot pose estimation
  via self-supervised sim-to-real transfer,'' in {\em Proceedings of the
  IEEE/CVF Conference on Computer Vision and Pattern Recognition},
  pp.~21296--21306, 2023.

\bibitem{labbe2021robopose}
Y.~Labb{\'e}, J.~Carpentier, M.~Aubry, and J.~Sivic, ``Single-view robot pose
  and joint angle estimation via render \& compare,'' in {\em Proceedings of
  the IEEE/CVF Conference on Computer Vision and Pattern Recognition},
  pp.~1654--1663, 2021.

\bibitem{lu2023}
J.~Lu, F.~Liu, C.~Girerd, and M.~C. Yip, ``Image-based pose estimation and
  shape reconstruction for robot manipulators and soft, continuum robots via
  differentiable rendering,'' in {\em 2023 IEEE International Conference on
  Robotics and Automation (ICRA)}, pp.~560--567, 2023.

\bibitem{milella2006computer}
A.~Milella, G.~Reina, R.~Siegwart, {\em et~al.}, ``Computer vision methods for
  improved mobile robot state estimation in challenging terrains.,'' {\em J.
  Multim.}, vol.~1, no.~7, pp.~49--61, 2006.

\bibitem{goldberg2003maximizing}
D.~Goldberg and M.~J. Matari{\'c}, ``Maximizing reward in a non-stationary
  mobile robot environment,'' {\em Autonomous Agents and Multi-Agent Systems},
  vol.~6, pp.~287--316, 2003.

\bibitem{zouaghi2011probabilistic}
L.~Zouaghi, A.~Alexopoulos, A.~Wagner, and E.~Badreddin, ``Probabilistic
  online-generated monitoring models for mobile robot navigation using modified
  petri net,'' in {\em 2011 15th International Conference on Advanced Robotics
  (ICAR)}, pp.~594--599, IEEE, 2011.

\bibitem{colle2019robust}
E.~Colle and S.~Galerne, ``A robust set approach for mobile robot localization
  in ambient environment,'' {\em Autonomous Robots}, vol.~43, pp.~557--573,
  2019.

\bibitem{rollinson2011state}
D.~Rollinson, A.~Buchan, and H.~Choset, ``State estimation for snake robots,''
  in {\em 2011 IEEE/RSJ International Conference on Intelligent Robots and
  Systems}, pp.~1075--1080, IEEE, 2011.

\bibitem{rollinson2013robust}
D.~Rollinson, H.~Choset, and S.~Tully, ``Robust state estimation with redundant
  proprioceptive sensors,'' in {\em Dynamic Systems and Control Conference},
  vol.~56147, p.~V003T40A005, American Society of Mechanical Engineers, 2013.

\bibitem{kalman1961new}
R.~E. Kalman and R.~S. Bucy, ``New results in linear filtering and prediction
  theory,'' 1961.

\bibitem{van2004sigma}
R.~Van Der~Merwe, E.~A. Wan, S.~Julier, {\em et~al.}, ``Sigma-point kalman
  filters for nonlinear estimation and sensor-fusion: Applications to
  integrated navigation,'' in {\em Proceedings of the AIAA guidance, navigation
  \& control conference}, vol.~3, p.~08, Providence, RI Providence, RI, 2004.

\bibitem{kirillov2023segment}
A.~Kirillov, E.~Mintun, N.~Ravi, H.~Mao, C.~Rolland, L.~Gustafson, T.~Xiao,
  S.~Whitehead, A.~C. Berg, W.-Y. Lo, {\em et~al.}, ``Segment anything,'' {\em
  arXiv preprint arXiv:2304.02643}, 2023.

\bibitem{chen2018encoder}
L.~C. Chen, Y.~Zhu, G.~Papandreou, F.~Schroff, and H.~Adam, ``Encoder-decoder
  with atrous separable convolution for semantic image segmentation,'' in {\em
  Proc. Europ. Conf. Comput. Vis.}, pp.~801--818, 2018.

\bibitem{chaumette2004image}
F.~Chaumette, ``Image moments: a general and useful set of features for visual
  servoing,'' {\em IEEE Transactions on Robotics}, vol.~20, no.~4,
  pp.~713--723, 2004.

\bibitem{schreiber2020arcsnake}
D.~A. Schreiber, F.~Richter, A.~Bilan, P.~V. Gavrilov, H.~M. Lam, C.~H. Price,
  K.~C. Carpenter, and M.~C. Yip, ``Arcsnake: An archimedes’ screw-propelled,
  reconfigurable serpentine robot for complex environments,'' in {\em 2020 IEEE
  International Conference on Robotics and Automation (ICRA)}, pp.~7029--7034,
  IEEE, 2020.

\bibitem{denavit1955kinematic}
J.~Denavit and R.~S. Hartenberg, ``A kinematic notation for lower-pair
  mechanisms based on matrices,'' 1955.

\bibitem{pytorch3d}
N.~Ravi, J.~Reizenstein, D.~Novotny, T.~Gordon, W.-Y. Lo, J.~Johnson, and
  G.~Gkioxari, ``Accelerating 3d deep learning with pytorch3d,'' {\em arXiv
  preprint arXiv:2007.08501}, 2020.

\bibitem{paszke2017automatic}
A.~Paszke, S.~Gross, S.~Chintala, G.~Chanan, E.~Yang, Z.~DeVito, Z.~Lin,
  A.~Desmaison, L.~Antiga, and A.~Lerer, ``Automatic differentiation in
  pytorch,'' 2017.

\bibitem{yao2020image}
G.~Yao, R.~Saltus, and A.~Dani, ``Image moment-based extended object tracking
  for complex motions,'' {\em IEEE Sensors Journal}, vol.~20, no.~12,
  pp.~6560--6572, 2020.

\bibitem{kingma2014adam}
D.~P. Kingma and J.~Ba, ``Adam: A method for stochastic optimization,'' {\em
  arXiv preprint arXiv:1412.6980}, 2014.

\end{thebibliography}

\end{document}